\documentclass{article}
\usepackage{spconf,amsmath,graphicx,hyperref}
\usepackage{etoolbox}
\usepackage{multirow}
\usepackage{adjustbox}
\usepackage{pifont}
\usepackage{amssymb}
\usepackage{booktabs}

\newcommand{\affmark}[1]{\textsuperscript{#1}}

\title{pFedSAM: Personalized Federated Learning of Segment Anything Model for Medical Image Segmentation}

\name{%
\begin{tabular}{c}
Tong Wang\affmark{1}\thanks{Equal contributions.}\quad
Xingyue Zhao\affmark{2}\footnotemark[1]\quad
Linghao Zhuang\affmark{1}\footnotemark[1]\quad
Haoyu Zhao\affmark{3}\\[2pt]
Jiayi Yin\affmark{4}\quad
Yuyang He\affmark{4}\quad
Gang Yu\affmark{5}\thanks{Corresponding author.}\quad
Bo Lin\affmark{4}\thanks{Corresponding author.}
\end{tabular}
}

\address{%
  \affmark{1} College of Computer Science and Technology, Zhejiang University, China\\
  \affmark{2} Chinese Academy of Medical Sciences and Peking Union Medical College, China\\
  \affmark{3} School of Computer Science, Wuhan University, China\\
  \affmark{4} Innovation Centre for Information, Binjiang Institute of Zhejiang University\\
  \affmark{5} Department of Data and Information, Children's Hospital, Zhejiang University School of Medicine\\[2pt]
  \textit{Corresponding emails:} yugang@zju.edu.cn,\; rainbowlin@zju.edu.cn
}

\begin{document}
\ninept
\maketitle
\begin{abstract}
Medical image segmentation is crucial for computer-aided diagnosis, yet privacy constraints hinder data sharing across institutions. Federated learning addresses this limitation, but existing approaches often rely on lightweight architectures that struggle with complex, heterogeneous data. Recently, the Segment Anything Model (SAM) has shown outstanding segmentation capabilities; however, its massive encoder poses significant challenges in federated settings. In this work, we present the first personalized federated SAM framework tailored for heterogeneous data scenarios in medical image segmentation. Our framework integrates two key innovations: (1) a personalized strategy that aggregates only the global parameters to capture cross-client commonalities while retaining the designed L-MoE (Localized Mixture-of-Experts) component to preserve domain-specific features; and (2) a decoupled global-local fine-tuning mechanism that leverages a teacher-student paradigm via knowledge distillation to bridge the gap between the global shared model and the personalized local models, thereby mitigating overgeneralization. Extensive experiments on two public datasets validate that our approach significantly improves segmentation performance, achieves robust cross-domain adaptation, and reduces communication overhead.
\end{abstract}
\section{Introduction}
Medical image segmentation is crucial for clinical decision-making but requires large-scale data, which is hindered by privacy concerns. Federated learning \cite{mcmahan2017communication} offers a solution by enabling collaborative training across institutions. However, medical images exhibit significant heterogeneity due to differences in acquisition conditions, equipment, and patient demographics. This phenomenon, known as domain drift, has motivated the development of Personalized Federated Learning.\cite{li2022federated,arivazhagan2019federated,t2020personalized,li2021fedbn,chen2021bridging,li2021ditto,yang2024fedas}. Moreover, existing segmentation methods, mostly based on lightweight architectures like U-Net \cite{ronneberger2015u}, struggle to handle such complex cross-domain variations.

\begin{figure}[t]
	\centering
	\includegraphics[width=\linewidth]{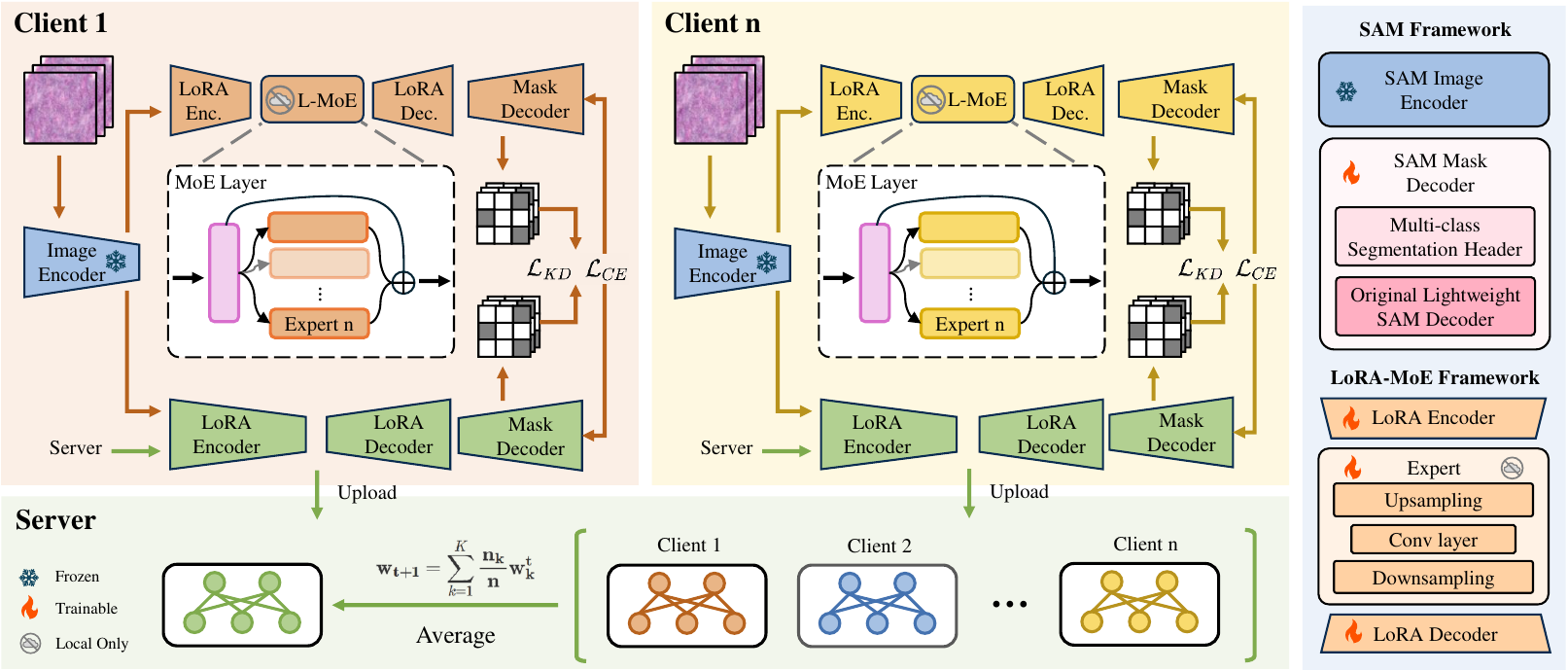} 
	\caption{\small Overview of the proposed personalized federated SAM framework. Left: Process flow of our framework. Right: Detailed architecture of SAM and the L-MoE structure.}
	\label{method}
\end{figure}

In contrast, the Segment Anything Model (SAM) \cite{kirillov2023segment,wu2023medical,le2024medficientsam, zhao2024sam} exhibits outstanding feature extraction and segmentation capabilities as a large-scale foundation model. However, its enormous encoder renders direct federated training and aggregation impractical. Although recent studies \cite{liu2024fedfms} have implemented SAM-based federated architectures with promising results, they universally overlook the critical need for personalization in medical image segmentation. This presents a critical challenge: \textit{how can we effectively harness SAM's strengths in federated settings while tailoring the model to the domain-specific characteristics of local datasets?}

To address these challenges, we propose a novel framework with two key innovations. \textbf{Firstly}, we achieve parameter-efficient adaptation and domain feature personalization by introducing a personalization strategy that decomposes the adaptable parameters into a Low-Rank Adaptation (LoRA)\cite{hu2022lora} module and a L-MoE component. Given SAM’s massive encoder, we employ LoRA for lightweight fine-tuning, which substantially reduces communication and computational costs. While LoRA is parameter-efficient, its limited expressiveness for heterogeneous data motivates the integration of a MoE design\cite{zhong2024convolution} to better capture domain-specific features. More critically, a fundamental challenge arises in the federated context: straightforwardly aggregating all parameters of such an architecture combining LoRA and L-MoE risks diluting the client-specific knowledge they are meant to preserve. Inspired by FedBN\cite{li2021fedbn}, we propose the L-MoE design to keep expert components local, as different experts specialize in distinct feature patterns unique to each client's data distribution. Thus, we propose a novel parameter-decomposing personalization strategy: only the global LoRA parameters, designed to capture cross-client commonalities, are uploaded and aggregated, while the L-MoE component, which preserve local domain characteristics, are retained on each client. This design leverages SAM's robust feature extraction while preventing domain drift, thereby enhancing both adaptation efficiency and segmentation performance on local data. 

\textbf{Secondly}, while the L-MoE design enables local personalization for each client, a conflict arises during the local update phase between the global model's pursuit of shared commonalities and the L-MoE component's specialization in client-specific features. Aggregating specialized L-MoE parameters from diverse local datasets is conceptually problematic, as it yields an ineffective "average expert" that dilutes the unique knowledge it is designed to preserve. To address this challenge, we propose an asymmetric federated learning framework that employs heterogeneous model architectures. In this framework, we decouple the roles of the global and local models. The global model, comprising only lightweight LoRA parameters, is responsible for learning shared commonalities with high communication efficiency. In contrast, each local model maintains a more expressive LoRA and L-MoE architecture specialized for its complex local data. To facilitate knowledge transfer between these disparate architectures, we introduce a distillation-based local update mechanism. By using the global LoRA-based SAM model as the teacher, we obtain smoother soft target distributions, which facilitates more effective transfer of global contextual patterns into personalized local models. 

Our main contributions are summarized as follows: 

(1) We propose a novel personalized federated SAM framework, named pFedSAM, tailored for heterogeneous data scenarios, adapting the powerful SAM model within a personalized federated learning setting.  

(2) We design a novel personalization strategy that decomposes adaptable parameters into a shared global component and local L-MoE modules, which are then harmonized using a decoupled knowledge distillation mechanism to ensure effective knowledge transfer across the resulting heterogeneous architectures.

(4) Our results demonstrate that pFedSAM achieves significantly improved segmentation performance and robust cross-domain adaptation on heterogeneous medical datasets.

\section{Method}
\subsection{Model Architecture}


We adapt the prompt-free SAM architecture, which comprises a ViT image encoder, a prompt encoder, and a mask decoder. As shown in Fig.~\ref{method}, we modify the image encoder by injecting trainable low-rank LoRA matrices into its transformer layers. Our LoRA-MoE variant extends this by inserting a gating network and convolutional experts between the LoRA matrices. The gating mechanism dynamically selects these experts, which utilize multi-scale up/downsampling operations to capture local priors.

\subsection{Overview of pFedPMS}

As shown in Fig.~\ref{method}, each client maintains a personalized local model and a shared global model. To balance personalization with parameter efficiency, we augment the local model's encoder with both L-MoE and LoRA modules, whereas the global model's encoder is adapted using only LoRA. The training process comprises three stages: local model training, global model training, and global aggregation.

For personalized training, we employ knowledge distillation instead of common regularization techniques~\cite{li2021ditto,xie2024perada,yang2024fedas}. The global LoRA model acts as a frozen teacher, providing guidance to the local L-MoE student model as it trains on local data. This process aligns local and global representations. The personalized model loss function $\mathcal{L}_{per}$ is represented as follows:
\begin{equation}
    \mathcal{L}_{per}=\mathcal{L}_{CE}\left(f\left(x,\theta_p\right), y\right)+\lambda_{L-MoE} \mathcal{L}_{L-MoE} + \lambda_{KD}\mathcal{L}_{KD}
\label{eq:loss}
\end{equation}
where $\theta_p$ is the personalized model parameters. $f\left(x,\theta_p\right)$ is the embedding output of $\theta_p$ and $y$ is the label of input data $x$ in $D_i$. $\mathcal{L}_{CE}$ is cross-entropy loss, while $\mathcal{L}_{L-MoE}$ and $\mathcal{L}_{KD}$ represent the loss functions of the L-MoE component and the knowledge distillation. The weight parameters $\lambda_{L-MoE}$ and $\lambda_{KD}$ are set to 1.5 and 0.1. Further details are provided in \ref{kd&MoE}.


During global model training, the global SAM model is trained independently on each client's dataset to acquire local knowledge. Finally, the server aggregates the collected global models via weighted averaging \cite{mcmahan2017communication} and distributes the updated model to clients for the next training round.

\begin{table*}[!htb] 
\centering
\caption{Comparison of different federated learning Methods on fundus datasets.} 
\label{tab:performance1}
\setlength{\tabcolsep}{4pt} 
\small 
\renewcommand{\arraystretch}{1.2}
    \begin{tabular}{l|cc|cc|cc|cc|cc|cc|cc}
    \toprule 
    \multicolumn{1}{c|}{Dataset} & \multicolumn{14}{c}{Prostate Cancer} \\
    \cmidrule(r){2-15} 
    \multicolumn{1}{c|}{Client} & \multicolumn{2}{c|}{HK} & \multicolumn{2}{c|}{I2CVB} & \multicolumn{2}{c|}{ISBI} & \multicolumn{2}{c|}{ISBI 1.5} & \multicolumn{2}{c|}{UCL} & \multicolumn{2}{c|}{Average} & \multicolumn{2}{c}{BIDMC (Unseen)} \\
    \cmidrule(r){1-1} \cmidrule(r){2-3} \cmidrule(r){4-5} \cmidrule(r){6-7} \cmidrule(r){8-9} \cmidrule(r){10-11} \cmidrule(r){12-13} \cmidrule(r){14-15}
    Model & Dice & IoU & Dice & IoU & Dice & IoU & Dice & IoU & Dice & IoU & Dice & IoU & Dice & IoU \\
    \midrule 
    FedSAM & 0.795 & 0.733 & 0.852 & 0.820 & 0.787 & 0.724 & 0.784 & 0.745 & 0.727 & 0.694 & 0.789 & 0.743 & 0.647 & 0.607 \\
    FedMSA & 0.751 & 0.705 & 0.793 & 0.806 & 0.749 & 0.664 & 0.798 & 0.752 & \textbf{0.736} & \textbf{0.708} & 0.765 & 0.727 & 0.638 & 0.615 \\
    Ditto & 0.782 & 0.714 & 0.799 & 0.763 & 0.756 & 0.708 & 0.811 & 0.768 & 0.702 & 0.651 & 0.770 & 0.721 & 0.765 & 0.724 \\
    FedBN & 0.797 & 0.752 & 0.826 & 0.844 & 0.761 & 0.705 & 0.803 & 0.765 & 0.713 & 0.657 & 0.780 & 0.745 & 0.768 & 0.717 \\
    FedPer & 0.844 & \textbf{0.809} & 0.879 & 0.852 & 0.773 & 0.718 & 0.756 & 0.715 & 0.722 & 0.661 & 0.795 & 0.751 & 0.770 & 0.720 \\
    FedAS & \textbf{0.847} & 0.788 & 0.886 & 0.858 & 0.758 & 0.700 & \textbf{0.815} & \textbf{0.775} & 0.714 & 0.656 & 0.804 & 0.755 & 0.577 & 0.551 \\
    \midrule 
    Ours & 0.833 & 0.772 & \textbf{0.898} & \textbf{0.869} & \textbf{0.790} & \textbf{0.735} & 0.812 & 0.771 & 0.730 & 0.669 & \textbf{0.812} & \textbf{0.763} & \textbf{0.774} & \textbf{0.737} \\
    \bottomrule 
    \end{tabular}
\end{table*}

\subsection{Knowledge Distillation across Heterogeneous Models}
\label{kd&MoE}

Given a pre-trained weight matrix $W_0$ with dimensions $d\times k$, the forward pass result based on LoRA can be defined as:
\begin{equation}
h=W_0x+BAx
\label{eq:lora}
\end{equation}
where $A$ and $B$ are low-rank decomposition matrices with dimensions $r\times k$ and $d\times r$, respectively, and $x$ is the input vector. $r$ is significantly smaller than the minimum of $d$ and $k$. L-MoE transforms eq.(\ref{eq:lora}) into:
\begin{equation}
    h=W_0x+B\left(\sum_i^m G\left(Ax\right) E_i\left(Ax\right)\right)
\end{equation}
where $G$ represents the gating mechanism and $E_i$ represents the $i_{t h}$ expert among $m$ available experts. The image encoder's forward propagation not only generates feature embeddings but also computes the MoE loss $\mathcal{L}_{L-MoE}$.

During the training process, we define the global model based on LoRA as the teacher and the personalized model based on LoRA and L-MoE as the student. Employing a LoRA-based global SAM model as the teacher network facilitates smoother soft target distributions throughout the distillation process. The knowledge distillation loss $\mathcal{L}_{DK}$ is as follows:
\begin{equation}
\mathcal{L}_{DK}=-\frac{\tau^2}{N} \sum_{i=1}^N\left[p_{t}^i \log \sigma\left(p_{s}^i\right)+\left(1-p_{t}^{i}\right) \log \left(1-\sigma\left(p_{s}^{i}\right)\right]\right.
\end{equation}
where $\tau$ is distillation temperature with default value 0.5 and $N$ is the product of the batch size, the image height, and the image width. $p_{s}^{i}$ is the probability distribution of the student after temperature scaling. Following temperature scaling, the teacher output is processed through a sigmoid activation function to derive $p_{t}^{i}$. The final loss function of personalized SAM model combines three components, as shown in eq.(\ref{eq:loss}).




\begin{table*}[!htb]
\centering
\caption{Comparison of different federated learning methods on Fundus datasets for OD segmentation.}
\label{tab:performance_fundus_od}
\small 
\setlength{\tabcolsep}{5pt} 
\renewcommand{\arraystretch}{1.2}

\begin{tabular}{@{}l|cc|cc|cc|cc|cc@{}}
\toprule
\multirow{2}{*}{Model} & \multicolumn{2}{c|}{ORIGA} & \multicolumn{2}{c|}{G1020} & \multicolumn{2}{c|}{Drishit-GS1} & \multicolumn{2}{c|}{Average} & \multicolumn{2}{c}{REFUGE (Unseen)} \\
\cmidrule(l){2-11}
 & Dice & IoU & Dice & IoU & Dice & IoU & Dice & IoU & Dice & IoU \\
\midrule
FedSAM & 0.886 & 0.800 & 0.890 & 0.812 & 0.759 & 0.618 & 0.845 & 0.743 & 0.875 & \textbf{0.783} \\
FedMSA & \textbf{0.888} & \textbf{0.801} & \textbf{0.900} & \textbf{0.821} & 0.734 & 0.591 & 0.841 & 0.738 & 0.856 & 0.754 \\
Ditto & 0.871 & 0.776 & 0.889 & 0.804 & 0.734 & 0.589 & 0.832 & 0.723 & 0.871 & 0.775 \\
FedBN & 0.880 & 0.790 & \textbf{0.900} & 0.810 & 0.830 & 0.720 & 0.869 & 0.774 & 0.876 & 0.774 \\
FedPer & 0.880 & 0.792 & 0.897 & 0.816 & 0.849 & 0.740 & 0.875 & 0.783 & 0.874 & 0.781 \\
FedAS & 0.875 & 0.784 & 0.891 & 0.808 & 0.857 & 0.753 & 0.874 & 0.782 & 0.872 & 0.768 \\
\midrule
\textbf{Ours} & 0.884 & 0.796 & 0.896 & 0.815 & \textbf{0.864} & \textbf{0.764} & \textbf{0.881} & \textbf{0.792} & \textbf{0.883} & 0.778 \\
\bottomrule
\end{tabular}

\end{table*}

\section{Experiment}

\subsection{Experimental Setup}

\subsubsection{Datasets}
We validate pFedSAM on publicly available medical image segmentation datasets: fundus datasets and prostate datasets. Fundus datasets contain four distinct fundus photography images datasets~\cite{bajwa2020g1020,orlando2020refuge,sivaswamy2015comprehensive,zhang2010origa}, targeting optic disc (OD) segmentation tasks. Prostate datasets consist of six datasets established from \cite{lemaitre2015computer,litjens2014evaluation} and NCI-ISBI 2013. In alignment with \cite{liu2024fedfms,wu2023medical}, all images in the datasets are resized to 1024 × 1024 × 3 during preprocessing, and the output mask is configured to 256 × 256. The transformation method for converting 3D prostate cancer images to 2D images is consistent with the approach described in \cite{liu2024fedfms}.

\begin{table*}[!htbp]
\centering
\caption{Ablation studies of pFedSAM.}
\label{tab:ablation_simplified}
\small 
\setlength{\tabcolsep}{4pt} 
\renewcommand{\arraystretch}{1.2}

\begin{tabular}{@{}lcccccccccc@{}} 
\toprule
\multirow{3}{*}{Method} & \multirow{3}{*}{Personalization} & \multicolumn{2}{c}{Client Model Arch.} & \multicolumn{2}{c}{Global Model Arch.} & \multirow{3}{*}{Distillation} & \multicolumn{2}{c}{Prostate} & \multicolumn{2}{c}{Fundus (OD)} \\
\cmidrule(lr){3-4} \cmidrule(lr){5-6} \cmidrule(lr){8-9} \cmidrule(l){10-11}
& & \multirow{2}{*}{LoRA} & \multirow{2}{*}{LoRA-MoE} & \multirow{2}{*}{LoRA} & \multirow{2}{*}{LoRA-MoE} & & \multirow{2}{*}{Dice} & \multirow{2}{*}{IoU} & \multirow{2}{*}{Dice} & \multirow{2}{*}{IoU} \\
& & & & & & & & & & \\
\midrule
Method A & & \checkmark & & \checkmark & & & 0.647 & 0.607 & 0.815 & 0.705 \\
Method B & & & \checkmark & & \checkmark & & 0.766 & 0.724 & 0.823 & 0.720 \\
Method C & \checkmark & & \checkmark & & \checkmark & & 0.787 & 0.723 & 0.877 & 0.785 \\
Method D & \checkmark & \checkmark & & \checkmark & & \checkmark & 0.764 & 0.715 & 0.871 & 0.777 \\
Method E & \checkmark & & \checkmark & & \checkmark & \checkmark & 0.773 & 0.731 & 0.876 & 0.785 \\
\midrule
\textbf{Ours} & \checkmark & & \checkmark & \checkmark & & \checkmark & \textbf{0.812} & \textbf{0.763} & \textbf{0.881} & \textbf{0.792} \\
\bottomrule
\end{tabular}

\end{table*}

\begin{table}[!htbp]
\centering
\small 
\caption{FLOPS and parameters of transfer model.}
\label{tab:flops}
\renewcommand{\arraystretch}{1.2}
\begin{tabular}{@{}llcc@{}} 
\toprule
Dataset & Model & FLOPS & Params \\
\midrule
\multirow{5}{*}{Fundus} & SAM & 1142G & 93.735M \\
 & SAM Adapter & \textbf{518G} & 14.7M \\
 & SAM LoRA & 820G & \textbf{4.212M} \\
 & SAM MoE & 822G & 4.226M \\
 & \textbf{Ours} & 820G & \textbf{4.212M} \\
\bottomrule
\end{tabular}
\end{table}

\subsubsection{Implementation Details}

We treat each dataset as a distinct client and hold one client per task as an unseen domain for generalization testing. Data from participating clients is split 90\%/10\% for training and testing. Performance is evaluated using the Dice coefficient and IoU. All models are implemented in PyTorch and trained on NVIDIA A800 GPUs with the following hyperparameters: an Adam optimizer ($\beta_1=0.9, \beta_2=0.999$), a batch size of 4, 4 L-MoE experts, a LoRA rank ($r$) of 4, and a scaling factor $\alpha$ of 16.

\subsection{Results and Discussion}

We conduct experimental comparisons with six federated learning approaches for SAM models, comprising two existing methods, FedSAM and FedMSA \cite{liu2024fedfms}, and four state-of-the-art personalized federated learning (pFL) approaches, Ditto \cite{li2021ditto}, FedAS \cite{yang2024fedas}, FedPer \cite{arivazhagan2019federated}, and FedBN \cite{li2021fedbn}, adapted for the SAM framework. These baselines include methods that achieve personalization through regularization (Ditto, FedAS) and those that designate specific layers for personalization to address domain drift (FedPer, FedBN). Due to the Non-independent and identically distributed (Non-IID) nature of the client datasets, achieving optimal performance across all domains is a significant challenge.

As shown in Table \ref{tab:performance1} and Table \ref{tab:performance_fundus_od}, our proposed method, pFedSAM, establishes a new state-of-the-art in average performance. Specifically, on the Prostate Cancer task, pFedSAM achieves an average Dice score of 0.812, surpassing all baselines. This superiority is mirrored in the Fundus task, where it leads with average Dice scores of 0.881 for OD. This consistent high performance across diverse tasks and clients underscores our model's advanced ability to mitigate domain drift and aggregate knowledge more effectively than existing pFL approaches in challenging Non-IID settings.

Furthermore, pFedSAM demonstrates stronger and more reliable generalization on unseen domain datasets. On the unseen BIDMC (Prostate) dataset, it significantly outperforms all competitors, creating a clear performance margin. While on the unseen REFUGE (Fundus) dataset, it achieves a highly competitive result, ranking among the top-tier methods. Collectively, these results validate that pFedSAM not only outperforms both regularization-based and personalized layer-based pFL methods in overall accuracy but also exhibits more robust generalization, making it a more effective solution for real-world federated medical imaging.

\subsubsection{Ablation Experiments}

To analyze the effectiveness of individual components in our method, we conducted extensive experiments to validate the impact of personalization of L-MoE, and heterogeneous distillation between the server and clients. Experimental results are presented in Table \ref{tab:ablation_simplified}. Model A and Model B represent the implementation of the standard method FedAvg \cite{mcmahan2017communication}. LoRA with MoE outperforms LoRA in federated learning of SAM. In Model C, we define L-MoE as a personalization layer constructed from the expert parameters of an MoE. Comparison between Model B and Model C suggests that personalization of MoE module is effective in domain drift problem. Model D and E outperform Model A and B indicates that knowledge distillation bridge the gap of the local personalization model and the global model. In the end, the experimental outcomes comparing our model with Models D and E demonstrate that adopting LoRA as the global teacher model ensures smoother soft distillation targets and prevents the local MoE-LoRA model from over-generalizing, thereby enhancing model performance.

\vspace{-9pt}
\subsubsection{FLOPS and Parameters}
Table \ref{tab:flops} reports FLOPs and parameters of the transfer models. pFedSAM keeps both compute and size low. Compared with full SAM (1142 G FLOPs, 93.74 M params), our LoRA-based model uses 820 G FLOPs and only 4.21 M params—about 28\% less compute and ~22× fewer parameters. Versus SAM-Adapter (518 G, 14.7 M), we spend more compute (~58\% higher) but cut parameters by ~3.5×, which reduces the amount sent each round and lowers memory on clients. Distillation between the LoRA and LoRA-MoE variants keeps performance while keeping the parameter count essentially unchanged (4.21 M vs. 4.23 M). Overall, we strike a practical balance: small to communicate, light to run, and strong enough to perform.

\section{Conclusion}


In this paper, we addressed the significant challenge of adapting SAM for federated medical image segmentation, particularly in the presence of client data heterogeneity. We introduced pFedSAM, a novel personalized federated learning framework that successfully integrates SAM's powerful capabilities into a privacy-preserving setting. Our approach features a parameter-decomposing strategy that separates global commonalities from domain-specific features, which are captured by a L-MoE component. Furthermore, we designed a decoupled global-local fine-tuning mechanism based on knowledge distillation to effectively transfer knowledge from the shared global model to the specialized local models. Our extensive experimental results on two public datasets demonstrate that pFedSAM not only achieves state-of-the-art segmentation performance but also exhibits strong cross-domain adaptability and reduced communication costs.

\clearpage



\bibliographystyle{splncs04}
\bibliography{references.bib}

\begin{thebibliography}{10}
\providecommand{\url}[1]{\texttt{#1}}
\providecommand{\urlprefix}{URL }
\providecommand{\doi}[1]{https://doi.org/#1}

\bibitem{arivazhagan2019federated}
Arivazhagan, M.G., Aggarwal, V., Singh, A.K., Choudhary, S.: Federated learning with personalization layers. arXiv preprint arXiv:1912.00818  (2019)

\bibitem{bajwa2020g1020}
Bajwa, M.N., Singh, G.A.P., Neumeier, W., Malik, M.I., Dengel, A., Ahmed, S.: G1020: A benchmark retinal fundus image dataset for computer-aided glaucoma detection. In: 2020 International Joint Conference on Neural Networks (IJCNN). pp.~1--7. IEEE (2020)

\bibitem{chen2021bridging}
Chen, H.Y., Chao, W.L.: On bridging generic and personalized federated learning for image classification. arXiv preprint arXiv:2107.00778  (2021)

\bibitem{hu2022lora}
Hu, E.J., Shen, Y., Wallis, P., Allen-Zhu, Z., Li, Y., Wang, S., Wang, L., Chen, W., et~al.: Lora: Low-rank adaptation of large language models. ICLR  \textbf{1}(2), ~3 (2022)

\bibitem{kirillov2023segment}
Kirillov, A., Mintun, E., Ravi, N., Mao, H., Rolland, C., Gustafson, L., Xiao, T., Whitehead, S., Berg, A.C., Lo, W.Y., et~al.: Segment anything. In: Proceedings of the IEEE/CVF international conference on computer vision. pp. 4015--4026 (2023)

\bibitem{le2024medficientsam}
Le, B.H., Nguyen-Vu, D.K., Nguyen-Mau, T.H., Nguyen, H.D., Tran, M.T.: Medficientsam: A robust medical segmentation model with optimized inference pipeline for limited clinical settings. In: CVPR 2024: Segment Anything In Medical Images On Laptop (2024)

\bibitem{lemaitre2015computer}
Lema{\^\i}tre, G., Mart{\'\i}, R., Freixenet, J., Vilanova, J.C., Walker, P.M., Meriaudeau, F.: Computer-aided detection and diagnosis for prostate cancer based on mono and multi-parametric mri: a review. Computers in biology and medicine  \textbf{60},  8--31 (2015)

\bibitem{li2022federated}
Li, Q., Diao, Y., Chen, Q., He, B.: Federated learning on non-iid data silos: An experimental study. In: 2022 IEEE 38th international conference on data engineering (ICDE). pp. 965--978. IEEE (2022)

\bibitem{li2021ditto}
Li, T., Hu, S., Beirami, A., Smith, V.: Ditto: Fair and robust federated learning through personalization. In: International conference on machine learning. pp. 6357--6368. PMLR (2021)

\bibitem{li2021fedbn}
Li, X., Jiang, M., Zhang, X., Kamp, M., Dou, Q.: Fedbn: Federated learning on non-iid features via local batch normalization. arXiv preprint arXiv:2102.07623  (2021)

\bibitem{litjens2014evaluation}
Litjens, G., Toth, R., Van De~Ven, W., Hoeks, C., Kerkstra, S., Van~Ginneken, B., Vincent, G., Guillard, G., Birbeck, N., Zhang, J., et~al.: Evaluation of prostate segmentation algorithms for mri: the promise12 challenge. Medical image analysis  \textbf{18}(2),  359--373 (2014)

\bibitem{liu2024fedfms}
Liu, Y., Luo, G., Zhu, Y.: Fedfms: Exploring federated foundation models for medical image segmentation. In: International Conference on Medical Image Computing and Computer-Assisted Intervention. pp. 283--293. Springer (2024)

\bibitem{mcmahan2017communication}
McMahan, B., Moore, E., Ramage, D., Hampson, S., y~Arcas, B.A.: Communication-efficient learning of deep networks from decentralized data. In: Artificial intelligence and statistics. pp. 1273--1282. PMLR (2017)

\bibitem{orlando2020refuge}
Orlando, J.I., Fu, H., Breda, J.B., Van~Keer, K., Bathula, D.R., Diaz-Pinto, A., Fang, R., Heng, P.A., Kim, J., Lee, J., et~al.: Refuge challenge: A unified framework for evaluating automated methods for glaucoma assessment from fundus photographs. Medical image analysis  \textbf{59},  101570 (2020)

\bibitem{ronneberger2015u}
Ronneberger, O., Fischer, P., Brox, T.: U-net: Convolutional networks for biomedical image segmentation. In: Medical image computing and computer-assisted intervention--MICCAI 2015: 18th international conference, Munich, Germany, October 5-9, 2015, proceedings, part III 18. pp. 234--241. Springer (2015)

\bibitem{sivaswamy2015comprehensive}
Sivaswamy, J., Krishnadas, S., Chakravarty, A., Joshi, G., Tabish, A.S., et~al.: A comprehensive retinal image dataset for the assessment of glaucoma from the optic nerve head analysis. JSM Biomedical Imaging Data Papers  \textbf{2}(1), ~1004 (2015)

\bibitem{t2020personalized}
T~Dinh, C., Tran, N., Nguyen, J.: Personalized federated learning with moreau envelopes. Advances in neural information processing systems  \textbf{33},  21394--21405 (2020)

\bibitem{wu2023medical}
Wu, J., Ji, W., Liu, Y., Fu, H., Xu, M., Xu, Y., Jin, Y.: Medical sam adapter: Adapting segment anything model for medical image segmentation. arXiv preprint arXiv:2304.12620  (2023)

\bibitem{xie2024perada}
Xie, C., Huang, D.A., Chu, W., Xu, D., Xiao, C., Li, B., Anandkumar, A.: Perada: Parameter-efficient federated learning personalization with generalization guarantees. In: Proceedings of the IEEE/CVF Conference on Computer Vision and Pattern Recognition. pp. 23838--23848 (2024)

\bibitem{yang2024fedas}
Yang, X., Huang, W., Ye, M.: Fedas: Bridging inconsistency in personalized federated learning. In: Proceedings of the IEEE/CVF Conference on Computer Vision and Pattern Recognition. pp. 11986--11995 (2024)

\bibitem{zhang2010origa}
Zhang, Z., Yin, F., Liu, J., Wong, W., Tan, N., Lee, B., Cheng, J., Wong, T.: Origa: An online retinal fundus image database for glaucoma analysis and research. In: Proceedings of the 2010 Annual International Conference of the IEEE Engineering in Medicine and Biology. pp. 3065--3068

\bibitem{zhao2024sam}
Zhao, X., Li, P., Luo, X., Yang, M., Chang, S., Li, Z.: Sam-driven weakly supervised nodule segmentation with uncertainty-aware cross teaching. In: 2024 IEEE International Symposium on Biomedical Imaging (ISBI). pp.~1--5. IEEE (2024)

\bibitem{zhong2024convolution}
Zhong, Z., Tang, Z., He, T., Fang, H., Yuan, C.: Convolution meets lora: Parameter efficient finetuning for segment anything model. arXiv preprint arXiv:2401.17868  (2024)

\end{thebibliography}

\end{document}